\pdfoutput=1

\documentclass[11pt]{article}

\usepackage[final]{acl}

\usepackage{times}
\usepackage{latexsym}

\usepackage[T1]{fontenc}

\usepackage[utf8]{inputenc}

\usepackage{microtype}

\usepackage{inconsolata}

\usepackage{graphicx}
\usepackage{cleveref}
\usepackage{makecell}
\usepackage{multirow}
\usepackage{hyperref}
\usepackage{arydshln}

\title{Contrastive Learning Using Graph Embeddings for Domain Adaptation of Language Models in the Process Industry}

\author{
Anastasia Zhukova\textsuperscript{1}\thanks{Equal contribution.}, Jonas L{\"u}hrs\textsuperscript{1}\footnotemark[1], Christian E. Lobm{\"u}ller\textsuperscript{2}, Bela Gipp\textsuperscript{1} \\
\textsuperscript{1}University of G{\"o}ttingen, Göttingen, Germany \\
\textsuperscript{2}eschbach GmbH, Bad S{\"a}ckingen, Germany \\ 
\texttt{anastasia.zhukova@uni-goettingen.de, jonas.luehrs@stud.uni-goettingen.de,} \\ 
\texttt{ christian.matt@eschbach.com, gipp@uni-goettingen.de}\\
}

\begin{document}
\maketitle

\begin{abstract}
Recent trends in NLP utilize knowledge graphs (KGs) to enhance pretrained language models by incorporating additional knowledge from the graph structures to learn domain-specific terminology or relationships between documents that might otherwise be overlooked. This paper explores how SciNCL, a graph-aware neighborhood contrastive learning methodology originally designed for scientific publications, can be applied to the process industry domain, where text logs contain crucial information about daily operations and are often structured as sparse KGs.
Our experiments demonstrate that language models fine-tuned with triplets derived from graph embeddings (GE) outperform a state-of-the-art mE5-large text encoder by 9.8-14.3\% (5.45-7.96p) on the proprietary process industry text embedding benchmark (PITEB) while having 3 times fewer parameters.

\end{abstract}

\section{Introduction}

For several years, domain adaptation of language models (LMs) has followed three well-established approaches: fine-tuning pre-trained LMs using labeled data of domain-specific tasks \cite{devlin-etal-2019-bert, liu_roberta_2019}, training domain-specific LMs from scratch \cite{lee_biobert_2019, beltagy_scibert_2019, huang_clinicalbert_2019}, or domain-adaptive continual pretraining (DAPT) using smaller amounts of domain text data when labeled domain task-specific data is limited \cite{gururangan_dont_2020, strubell_energy_2020}. While DAPT and the subsequent fine-tuning have become a practical and efficient solution for creating specialized NLP applications \cite{guo_domain_2022}, it relies on the availability of unlabeled domain-specific data, from which an LM can learn domain-specific terminology. 

When domain-specific unlabeled text data is limited, self-supervised methods such as SPECTER suggest leveraging the underlying graph as an additional data source structure to enhance training on the passage or document level \cite{cohan_specter_2020}. Such a graph represents documents as nodes and their relationships as edges, providing valuable contextual information about how documents are interconnected \cite{kinney_semantic_2023}. Incorporating this graph-based information into an LM’s training process enables the model to capture inter-document relationships \cite{cohan_specter_2020}. 

\begin{figure}
    \centering
    \includegraphics[width=\linewidth]{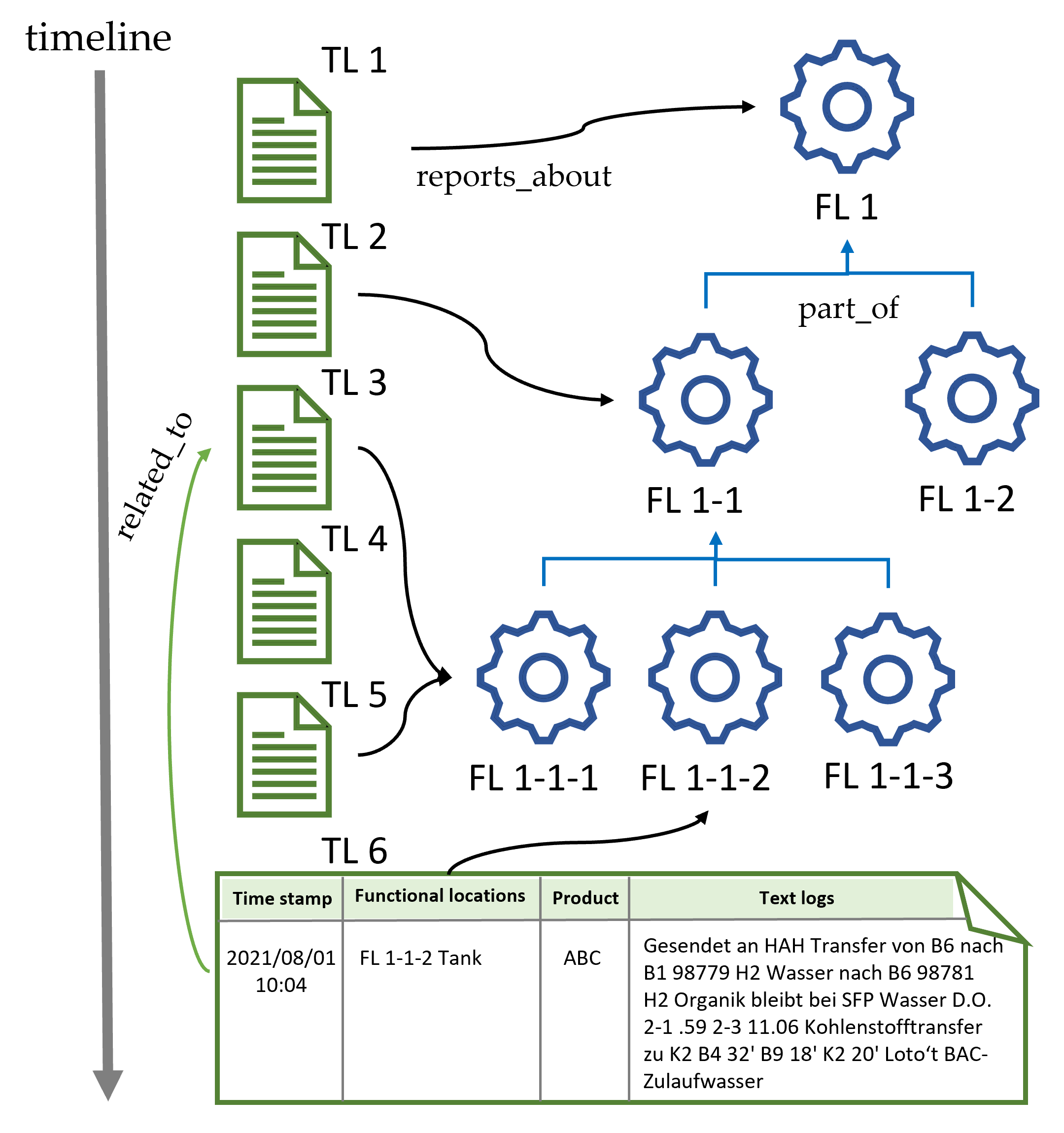}
    \caption{Graph embeddings are obtained from a directed heterogeneous domain graph for the process industry with two node types (1) text log (TL), i.e., logs of the daily operations at a production plant, and (2) functional locations (FL), i.e., hierarchically structured machinery on a production plant, and three edge types: \textit{related\_to} (green) connects two text logs, \textit{reports\_about} (black) links a text log to FLs, and \textit{part\_of} (blue) represents the hierarchical structure among FLs.
    }
    \label{fig:plant_graph}
\end{figure}

Domain-specific areas that involve proprietary knowledge in languages other than English often fall into the low-resource setting when both unlabeled and labeled text data are scarce \cite{usuga_cadavid_valuing_2020, zhong_natural_2024}. Although the text data can be scarce, companies often collect, store, and use structured and/or graph data, for example, those belonging to the process industry domain (\Cref{fig:plant_graph}). Text records in this domain are logs of daily plant operations, organized in shift books, which preserve valuable knowledge about production and maintenance \cite{may_applying_2022, Zhukova2024}. 

This paper explores adapting the SciNCL methodology, a contrastive learning approach based on scientific graph embeddings (GE) \cite{ostendorff_neighborhood_2022}, to enhance the text encoder in German plant operation logs and improve a semantic search task in this domain. The goal is to explore the potential of fine-tuning small models in low-resource settings, additionally aiming to achieve cost-efficient inference in production. The experiments show that the best-performing model was obtained by fine-tuning mBERT using triplets generated based on the GE. 

The primary objective of adapting SciNCL is to leverage the relationships within the knowledge graph (KG) to provide an additional signal indicating the semantic relatedness between text logs, particularly in terms of the terminology used, e.g., "FL 1-1" as an abbreviation and “\textit{Lömi}” as jargon. In example from \Cref{fig:plant_graph}, one text log (TL2) reports about a reactor with a short code "FL 1-1" and uses \textit{Lömi} alongside, and another text log (TL6) reports about its component with a short code "FL 1-1-2" but uses the textbook terminology of \textit{Lömi}, i.e., \textit{Lösungsmittel} (solvent). When using the KG and specifically graph embeddings, we can encode the information that these logs are connected via the chain of edges \textit{TL6 $\rightarrow$ reports\_about $\rightarrow$ FL 1-1-2 $\rightarrow$ part\_of $\rightarrow$ FL 1-1 $\leftarrow$ reports\_about $\leftarrow$ TL2}. Incorporating domain-specific semantic relationships that were not available at the time of LM pretraining aims at bringing jargons, such as \textit{Lömi-Lösungsmittel}, closer together in the vector space.  

Our evaluation showed that mBERT outperformed a state-of-the-art mE5-large bi-encoder by 14.3\% (7.96p) and the best-performing baseline M3 by 1.5\% (0.92p) on the domain process industry text embedding benchmark (PITEB) while requiring considerably less training data (2.45M text pairs) and having 3 times fewer parameters (179M vs 560M).

\section{Related work}

\begin{figure*}
    \centering
    \includegraphics[width=\textwidth]{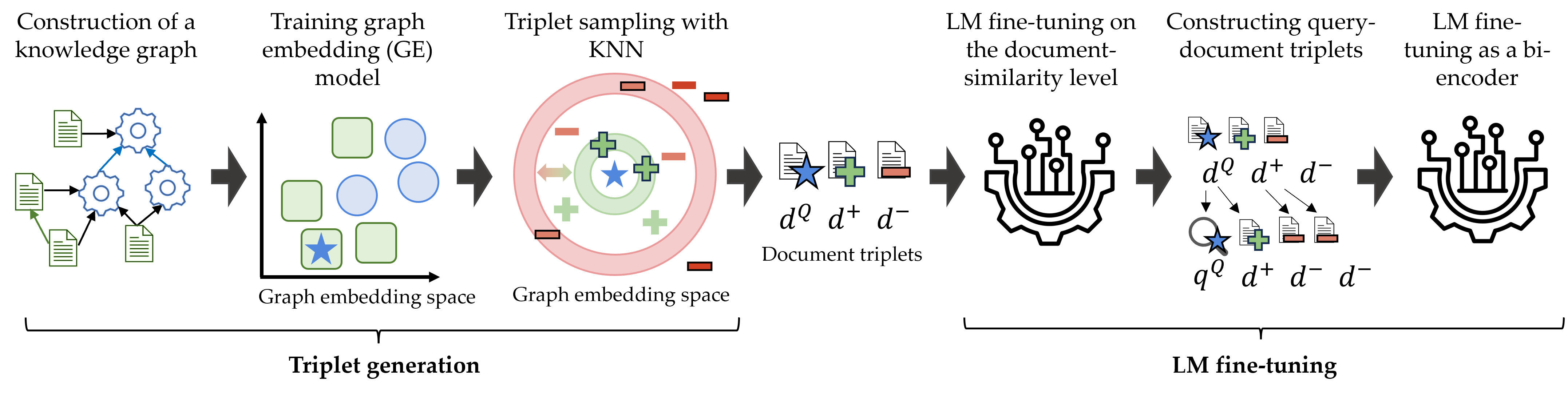}
    \caption{The methodology of adapting SciNCL \cite{ostendorff_neighborhood_2022} to a semantic search in the domain of the process industry. The two main changes involve generating document triplets using graph embeddings (GE) constructed from a heterogeneous knowledge graph (KG) and using these triplets as a source for query-document triplet generation during bi-encoder fine-tuning. }
    \label{fig:proposed_methodology}
\end{figure*}

A specific area of NLP, known as Technical Language Processing (TLP) \cite{brundage_technical_2021, may_applying_2022, akhbardeh_maintnet_2020}, adapts traditional NLP methods to address the unique challenges of the technical industry. The state-of-the-art NLP models often fall short when dealing with domain-specific technical terms, abbreviations \cite{akhbardeh_maintnet_2020}, incomplete sentences, typographical errors, and non-standard notations \cite{dima_adapting_2021}. 

Knowledge graphs (KGs) provide rich structural information and have been widely applied in various domains, including academic graphs, industrial manufacturing, and maintenance \cite{xia_maintenance_2023, xiao_knowledge_2023, stewart_mwo2kg_2022}. Knowledge embedding techniques transform these graphs into continuous embeddings for entities and their relationships \cite{wang_kepler_2021}. Since general-purpose language models lack domain-specific knowledge \cite{qiu_pre-trained_2020}, several approaches have been developed to enhance language representation by integrating information from external knowledge bases.  The recent development of the NLP applications in the process industry domain utilizes KGs for decision support \cite{naqvi_unlocking_2024}, predictive maintenance \cite{naqvi_leveraging_2022, usuga-cadavid_using_2022}, or semantic search \cite{naqvi_unlocking_2024, Zhukova2024}. 

Research on knowledge graph-based fine-tuning techniques includes methods such as K-BERT \cite{liu_k-bert_2020}, ERNIE \cite{zhang_ernie_2019}, KnowBERT \cite{peters_knowledge_2019}, and KEPLER \cite{wang_kepler_2021}, which directly incorporate entity embeddings and knowledge into the model's training process. Other studies focus on utilizing the graph structure of documents for learning document-level semantics, such as SPECTER \cite{cohan_specter_2020}, LinkBERT \cite{yasunaga_linkbert_2022}, and SciNCL \cite{ostendorff_neighborhood_2022}, which use citation information and document relations to fine-tune models. TwHIN-BERT \cite{zhang_twhin-bert_2023} applies contrastive learning on social media data, while MICoL \cite{zhang_metadata-induced_2022} uses metadata-induced contrastive learning. SciNCL, unlike other methods, utilizes graph embedding to sample both positive and negative documents, making it effective for adapting models to the process industry by learning from graph structures and incorporating hard-to-learn negatives \cite{bucher_hard_2016, wu_sampling_2017}.

\section{Methodology}
\label{sec:methodology}

We closely follow the implementation of SciNCL, with several modifications and adaptations tailored to the domain of the process industry. The following section describes the key changes to the SciNCL methodology. \Cref{sec:appendex_knowledge} until \Cref{sec_appendix_triplet} provides more details about each implementation stage.

\paragraph{Input data} A KG in the process industry differs from the scientific graph in passage length and graph connectivity. \Cref{tab:scientific_vs_process_industry} provides an overview of the key distinctions between the input data for the process industry and the original SciNCL approach. Unlike scientific abstracts, which maintain a consistent length and writing style to summarize research papers and have dense connectivity, text logs in the process industry are not subject to the writing constraints and result in a sparse graph. The text logs have a style and syntax similar to quickly taken notes (i.e., prone to typos), contain a lot of professional jargon and abbreviations, may contain partial information, are formatted inconsistently, and may exhibit other irregularities compared to commonly structured text. This results in logs as short as a few words, posing challenges for language learning objectives.

\begin{table}[]
\small
    \centering
    \begin{tabular}{ lll }
    \hline
        Params & SciNCL & PI-adapt. \\
        \hline
        Domain & scientific & proc. industry \\
        Document type & \makecell[l]{title + abstract} & text logs \\        
        \makecell[l]{Avg. passage\\ length (in words)} & 150-250 & 8-21 \\
        Graph connectivity & high & low \\
        Data quality & high & low \\  
        Graph type & homogeneous & heterogeneous \\
        \# nodes & >52M & 22K-172K \\      
        \# edges & >462M & 24K-1.8M \\
        \hline
        Training data & \makecell[l]{S2ORC \\\cite{lo_s2orc_2020} \\($\approx$82.5Gb)} & \makecell[l]{domain data \\(4 plants)\\ ($\approx$0.25Gb)}\\
        \hline
    \end{tabular}
    \caption{A summary table of the discrepancies of the original methodology of SciNCL to the proposed implementation adapted to the process industry. While small, these differences may significantly impact the effectiveness of leveraging domain KGs for language model fine-tuning.}
    \label{tab:scientific_vs_process_industry}
\end{table}

\paragraph{Knowledge graph} Our implementation expands the scientific graph from the homogeneous graph with one relation type to a heterogeneous graph (i.e., text logs and functional locations (FLs)) with three relation types: (1) \textit{part\_of} relation to represent the hierarchical structure between FLs, (2) \textit{reports\_about} relation to connect text logs to the FLs they reference, and (3) \textit{related\_to} relation to link two text logs when the second log is a follow-up event to the first, such as a solution to a reported problem. These edges are directed, with clearly defined source and target nodes. By combining these three relations, the graph effectively captures the domain-specific relationships inherent in process industry data. To improve graph connectivity, we perform a preprocessing step using link prediction with two custom models. First, we utilize named entity linking to restore \textit{reports\_about} relations, and then record linking to restore \textit{related\_to} relations \cite{zhukova2025linkpredictioneventlogs}.

\paragraph{Graph embeddings} A graph embedding (GE) model is designed to capture the domain-specific relationships within the process industry graph. During training, the model learns to position nodes connected by edges closer together in the embedding space. While node embeddings are typically initialized with random values, we improve the learning process by initializing the node embeddings with the sentence transformers applied to (1) text logs and (2) FL descriptions. This approach is similar to the method used by  \citet{asada_representing_2021}, who incorporated text embeddings to enhance the training of a graph embedding model in cases where the graph's structural information alone was insufficient.

\paragraph{Document triplet sampling} We employ the graph embedding neighborhood of a query document vector $\mathbf{d}^Q$ to sample positives and negatives by selecting the $k$-nearest neighbors. Graph embeddings provide a continuous and undirected similarity signal, enabling the identification of semantically similar nodes even when there are no direct edges between them. In a heterogeneous graph setup, node embeddings inherently encode the diverse contexts introduced by various relation types, thereby eliminating the need for complex sampling strategies that rely on directly building positive samples from graph edges \cite{zhang_metadata-induced_2022}. Although edges between FLs and text logs help define the graph embedding space, the resulting node embeddings ultimately become independent of these direct connections. We adopt the same sampling strategies for positives and negatives as SciNCL, employing kNN sampling for positives and hard negatives, and filtered random sampling for easy negatives. The sampling parameters were set as follows:  $c^+=2$ for positives, $c_{easy}^{-}=1$ for easy negatives, and $c_{hard}^{-}=1$ for hard negatives, where positives were selected given the kNN parameter $k^+=2$, and a hard negative was chosen from $k_{hard}^{-}=50$. 

\paragraph{Learning objective: document similarity level}
A contrastive learning objective ensures that similar documents are positioned close together in the embedding space of a fine-tuned LM while dissimilar documents are pushed farther apart. We fine-tune an LM with a self-supervised triplet margin loss \cite{schroff_facenet_2015}:
\begin{equation}
    \mathcal{L} = \max \left\{ \left\| \mathbf{d}^\mathcal{Q} - \mathbf{d}^+ \right\|_2 - \left\| \mathbf{d}^\mathcal{Q} - \mathbf{d}^- \right\|_2 + \xi, 0 \right\}
    \label{eq:triplet_loss}
\end{equation}

where $d$ is a document vector representation $f_{lm}(d) = \mathbf{d}$ using an LM, $\xi$ represents the margin that ensures that $\mathbf{d}^+$ is at least $\xi$ closer to $\mathbf{d}^\mathcal{Q}$ than $\mathbf{d}^-$, and $\|\Delta d\|_2$ is the $L^2$ norm, which is used as a distance function. In our experiments, we use $d$ as (1) a vector of \texttt{[CLS]} token from the last hidden layer or (2) a concatenation of \texttt{[CLS]} and mean pooling of all tokens from the last hidden layer. 

\paragraph{Query-document triplet sampling}
To create query-document pairs based on GE-triplets, we begin by generating a search query for each query document \cite{zhukova-etal-2025-automated}. Next, we apply a hard negative mining strategy to train Sentence Transformers using the MS MARCO dataset \cite{sentence_transformers_msmarco_hard_negatives}: we treat all documents retrieved by any model as negatives (both positive and negatives), except for the query document itself. As a result, all retrieved triplet documents are considered negatives, and previously positive documents are now treated as hard-to-learn negatives.

\paragraph{Bi-encoder fine-tuning}
We use Multiple Negatives Ranking Loss to train a bi-encoder from the implementation from the BEIR benchmark \cite{beir_2021}\footnote{\label{note1}\url{https://github.com/beir-cellar/beir/blob/main/examples/retrieval/training/train_msmarco_v3.py}.}.

\section{Evaluation}

\subsection{Experiment setup}
\paragraph{Model selection}
For LM fine-tuning, we selected several pre-trained and domain-adapted mono-lingual or multilingual pre-trained LMs. Specifically, we selected state-of-the-art LMs with BERT architectures that support the German language: GBERT-base \cite{chan_germans_2020}, multilingual mBERT-base \cite{devlin-etal-2019-bert}, and XLM-RoBERTa-base \cite{conneau-etal-2020-unsupervised}. Additionally, we evaluate daGBERT, a domain LM tailored for the process industry using continual pretraining \cite{Zhukova2025a}. All selected models have an embedding size of 768, and the number of parameters is under 300M.   

\paragraph{Baselines}
The evaluation compares our methodology to the state-of-the-art multilingual text encoders that support the German language: 
SBERT {\small\texttt{paraphrase-multilingual-MiniLM-L12-v}} and {\small\texttt{msmarco-distilbert-multilingual}} \cite{reimers_sentence-bert_2019}, mE5-base and mE5-large \cite{wang2024multilinguale5textembeddings}, IBM Granite 107m and 278m \cite{granite2024embedding}, OpenAI \texttt{text-embedding-3-large} \cite{openai2024embedding}, mGTE-base \cite{zhang-etal-2024-mgte}, Nomic Embed v2 \cite{nussbaum2025trainingsparsemixtureexperts}, \texttt{deepset-mxbai-embed-de-large-v1} \cite{germanemb2024mxbai}, and M3 \cite{chen-etal-2024-m3}. The selected models vary in terms of embedding size and number of parameters, thereby ensuring a fair comparison in terms of performance and efficiency. Adding these models to our evaluation assesses their performance in highly specific, technical domains, such as the process industry and the German language, thereby contributing to the exploration of their limits/boundaries for such edge cases.

\paragraph{Implementation}
The models were trained on an NVIDIA Tesla A100 (80GB). Adam with weight decay was used as the optimizer, with a learning rate of $\lambda=2^{-5}$. Our preliminary experiments showed that the most optimal pooling strategy for GBERT, daGBERT, and mBERT is concatenation, whereas for XLM-RoBERTa, it is using the [CLS] vector token. The training was conducted for three epochs with a batch size of 16. We fine-tuned a GBERT, daGBERT, and mBERT for 3 epochs and XLM-RoBERTa for 5 epochs. We trained a bi-encoder for 5 epochs on GBERT, daGBERT, and mBERT, using a batch size of 64 and 1000 warm-up steps. In contrast, due to the larger number of training parameters, XLM-RoBERTa was fine-tuned for 8 epochs with 3000 warm-up steps with the same batch size. The remaining parameters were inherited from SciNCL.

\paragraph{Test collection: PITEB}
We evaluate our models and baselines using a private expert-verified process industry text embedding benchmark (PITEB) \cite{zhukova-etal-2025-automated}. \Cref{tab:sem_se} provides details on the test collection, which comprises data from seven production plants in the chemical and pharmaceutical domains, totaling 205 queries across a collection of 330K documents. Data from four plants is used to build training datasets for both document-level triplets and a dataset for bi-encoders. 

\begin{table}[]
\centering
\small
\begin{tabular}{lrrrc}
\hline
Plant & \# docs & \makecell{\# queries} & \makecell{\# docs \\relevant} &\makecell{GE-based\\ triplets} \\
\hline
A & 17K & 30 & 2266  & +\\
B & 14K & 30 & 1747  & - \\
C & 129K & 30 & 1698 & + \\
D & 71K & 30 & 3799  & +\\
E & 10K & 28 & 1097  & - \\
F & 26K & 28 & 1894  & - \\
G & 64K & 29 & 2179  & + \\
\hline
total & 330K & 205 & 14680 \\
\hline
\end{tabular}
\caption{Process industry text embedding benchmark (PITEB) for semantic search evaluation. The data from three plants was unseen during the LM fine-tuning on the document level.}
\label{tab:sem_se}
\end{table}

\begin{table}[]
\small
\centering
\begin{tabular}{lrrr}
\hline
Component & Pos.pairs & Neg.pairs & Total\\
\hline
\makecell[l]{domain-related\\ MS MARCO} & 132K & 2.14M & 2.27M \\
\hdashline
\makecell[l]{synthetic\\ domain data \\(plants A-F)} &  9K & 32K & 41K \\
\hdashline
\makecell[l]{GE-based \\ query-doc pairs \\(plants A, C, D, G)} &  20K & 80K & 100K \\
\hline
\makecell[c]{Total} & 161K & 2.25M & 2.41M \\
\hline
\end{tabular}
\caption{The query-document pair datasets used for bi-encoder fine-tuning. A combination of domain-related MS MARCO and synthetic domain data is labeled as \texttt{DR-MM + SID} dataset, and the combination of all three components is labeled as \texttt{DR-MM + SID + GET}.}
\label{tab:data}
\end{table}

\paragraph{Document similarity training dataset}
We generated 100K document triplets equally distributed from four plants (50K unique query documents). We have enforced an additional quality check of the triplets by following a Sentence Transformers methodology of creating hard negatives for the MS MARCO dataset\footref{note1}: we scored the pairs with a cross-encoder {\small\texttt{cross-encoder/msmarco-MiniLM-L12-en-de-v1}} and kept the triplets where a positive document is similar enough to a query-document ($pos.pair \ge5.0$) and a negative document is distant enough from the positive document ($pos.pair - neg.pair \ge 3.0$). Hence, we ensured a second quality layer after the GEs and obtained the final training dataset with 14K triplets.

\begin{table*}[]
\centering
\scriptsize
\begin{tabular}{l|c|c|l|lll|l}
\hline
Model & \makecell{Params,\\ M} & \makecell{Doc.-sim.\\fine-tuning} & \makecell{Bi-encoder\\fine-tuning} & MAP@10 & MRR@10 & nDCG@10 & Mean  \\
\hline
\href{https://huggingface.co/sentence-transformers/paraphrase-multilingual-MiniLM-L12-v2}{SBERT/MiniLM-L12-v2} & 118  & - & \makecell[c]{-} & 43.84 & 46.99 & 26.90 & 39.24 \\
\href{https://huggingface.co/sentence-transformers/msmarco-distilbert-multilingual-en-de-v2-tmp-lng-aligned}{SBERT/msmarco-distilbert-multilingual} & 135  & - & \makecell[c]{-} & 54.83 & 59.81 & 35.12 & 49.92  \\
\href{https://huggingface.co/intfloat/multilingual-e5-base}{intfloat/multilingual-e5-base}  & 278  & - & \makecell[c]{-} & 58.78 & 65.08 & 40.28 & 54.71  \\
\href{https://huggingface.co/intfloat/multilingual-e5-large}{intfloat/multilingual-e5-large}  & 560  & - & \makecell[c]{-} & 59.82 & 65.31 & 42.26 & 55.80   \\
\href{https://huggingface.co/ibm-granite/granite-embedding-107m-multilingual}{ibm-granite/granite-embedding-107m-multiling.}  & 107  & - & \makecell[c]{-} & 62.42 & 67.45 & 44.67 & 58.18  \\
\href{https://platform.openai.com/docs/models/text-embedding-3-large}{OpenAI-text-embedding-3-large} & UNK  & - & \makecell[c]{-} & 63.68 & 68.57 & 45.60 & 59.28  \\
\href{https://huggingface.co/Alibaba-NLP/gte-multilingual-base}{Alibaba-NLP/gte-multilingual-base}  & 305  & - & \makecell[c]{-} & 63.44 & 68.76 & 46.03 & 59.41  \\
\href{https://huggingface.co/nomic-ai/nomic-embed-text-v2-moe}{nomic-ai/nomic-embed-text-v2-moe} & 305  & - & \makecell[c]{-} & 62.20 & 68.70 & 47.99 & 59.63   \\
\href{https://huggingface.co/ibm-granite/granite-embedding-278m-multilingual}{ibm-granite/granite-embedding-278m-multiling.} & 278  & - & \makecell[c]{-} & 64.44 & 70.22 & 46.26 & 60.31  \\
\href{https://huggingface.co/mixedbread-ai/deepset-mxbai-embed-de-large-v1}{mixedbread-ai/deepset-mxbai-embed-de-large-v1} & 487  & - & \makecell[c]{-} & \underline{\textit{65.32}} & 70.33 & 48.13 & 61.26 \\
\href{https://huggingface.co/BAAI/bge-m3}{BAAI/bge-m3} & 560  & - & \makecell[c]{-} & \textbf{\textit{66.24}} & 71.33 & \textit{\textbf{50.94}} & \textbf{\textit{62.84}} \\
\hline
\multirow{4}{*}{GBERT-base \cite{chan_germans_2020} }  &  & - & MM & 59.86 & 65.72 & 42.99 & 56.19 \\
  & 111  & - & DR-MM + SID & 62.64 & 66.92 & 45.78 & 58.45  \\
 &   & - & DR-MM + SID + GET & 64.58 & \textbf{\textit{71.48}} & 50.42 & 62.16  \\
  &   & + & DR-MM + SID + GET & 64.08 & \underline{\textit{71.42}} & 49.97 & 61.82  \\
\hline
\multirow{3}{*}{daGBERT-base \cite{Zhukova2025a}} &   & - & DR-MM + SID & 62.21 & 69.36 & 45.81 & 59.13   \\
  & 111  & - & DR-MM + SID + GET & 62.36 & 69.24 & 48.06 & 59.89  \\
  &  & + & DR-MM + SID + GET & 64.60 & 69.24 & 48.06 & 61.24  \\
\hline
\multirow{3}{*}{mBERT \cite{devlin-etal-2019-bert}}  &   & - & DR-MM + SID & 64.81 & 70.85 & 48.11 & 61.26 \\
  & 179   & - & DR-MM + SID + GET & 65.22 & 70.86 & \underline{\textit{50.70}} & \underline{\textit{62.26}} \\
  &  & + & DR-MM + SID + GET & \textbf{67.12} & \textbf{72.58} & \textbf{51.56} & \textbf{63.75} \\
\hline
\multirow{3}{*}{XLM-RoBERTa \cite{conneau-etal-2020-unsupervised}} &   & - & DR-MM + SID & 62.51 & 67.29 & 46.10 & 58.63 \\
 & 278  & - & DR-MM + SID + GET & 64.47 & 69.88 & 48.82 & 61.06 \\
    &   & + & DR-MM + SID + GET & 64.18 & 70.23 & 49.35 & 61.25\\
\hline
\end{tabular}
\caption{The evaluation demonstrates that fine-tuning language models (LMs) using the adapted SciNCL methodology outperforms the baseline text encoders. The most significant improvement is observed with the two-stage model fine-tuning  (first on document-level similarity and then on query-document level) and the \texttt{DR-MM + SID + GET} dataset, which shows the highest impact across almost all fine-tuned models. This result is achieved despite the GE-based triplets comprising only 6\% of the entire training corpus.}
\label{tab:results}
\end{table*}

\paragraph{Bi-encoder training dataset}
The dataset for bi-encoder fine-tuning consists of three components (\Cref{tab:data}): (1) the domain-related version of German MS MARCO \cite{NguyenRSGTMD16} (\texttt{DR-MM}) (see \Cref{sec:appendix_dataset}), (2) a synthetic in-domain dataset created using an ensemble of text encoders (\texttt{SID}) \cite{zhukova-etal-2025-automated}, and (3) a collection of query-document pairs generated from GE-based document-level triplets (\texttt{GET}). To create \texttt{GET}, we sampled 5K query documents from plants A, C, D, and G from the unique 50K query documents from the triplets above, and ensured that these documents do not occur in the test collection as relevant documents. To evaluate the impact of the query-document pairs based on GE-triplets, we use two versions of the dataset: (1) \texttt{DR-MM + SID} (2.31M pairs), (2) \texttt{DR-MM + SID + GET} (2.41M pairs). As a baseline, we also use the German MS MARCO (\texttt{MM}, 160M pairs) as a comparison dataset for our domain-specific version.

\paragraph{Metrics} We evaluate the text encoder with three metrics: mean average precision (MAP@10), mean reciprocal rank (MRR@10), and normalized discounted cumulative gain (nDCG@10). Using all three metrics together provides a more well-rounded evaluation, giving insights into ranking quality, early retrieval performance, and overall retrieval quality. We report the mean of these metrics across seven plants.

\subsection{Results}
\Cref{tab:results} shows that fine-tuning LMs with GE-based query-document triplets, whether used alone or combined with preliminary fine-tuning at the document level, outperforms 9 out of 11 text encoder baselines. The best model performance was achieved by fine-tuning mBERT with a two-stage fine-tuning process (first on document-level similarity and then on query-document level), which outperformed the state-of-the-art mE5-large by 14.3\% (7.96 points) and the strongest baseline, M3, by 1.5\% (0.92 points). Despite having three times fewer parameters and a smaller embedding dimension, fine-tuned mBERT surpassed these baselines. 

The analysis of the bi-encoder dataset reveals a positive effect from using domain-related query-document pairs from MS MARCO (\texttt{DR-MM + SID}), particularly when combined with in-domain data from GE-based query-document pairs (\texttt{GET}). The results show that fine-tuning GBERT using \texttt{DR-MM + SID} is 4\% (2.27 points) more efficient than using the full German MS MARCO \texttt{MM}, while requiring significantly less training data (1.4\%). Although \texttt{DR-MM + SID} includes synthetic domain data, its contribution is minimal, as it makes up just 1.8\% of the dataset. The greatest performance improvement comes from the GE-based query-document pairs, which accounted for only 5.6\% of the \texttt{DR-MM + SID + GET} bi-encoder dataset, as this component improved model performance across all evaluated models by 1.28-6.35\% compared to training on \texttt{DR-MM + SID} alone. 

\begin{figure}
    \centering
    \includegraphics[width=\linewidth]{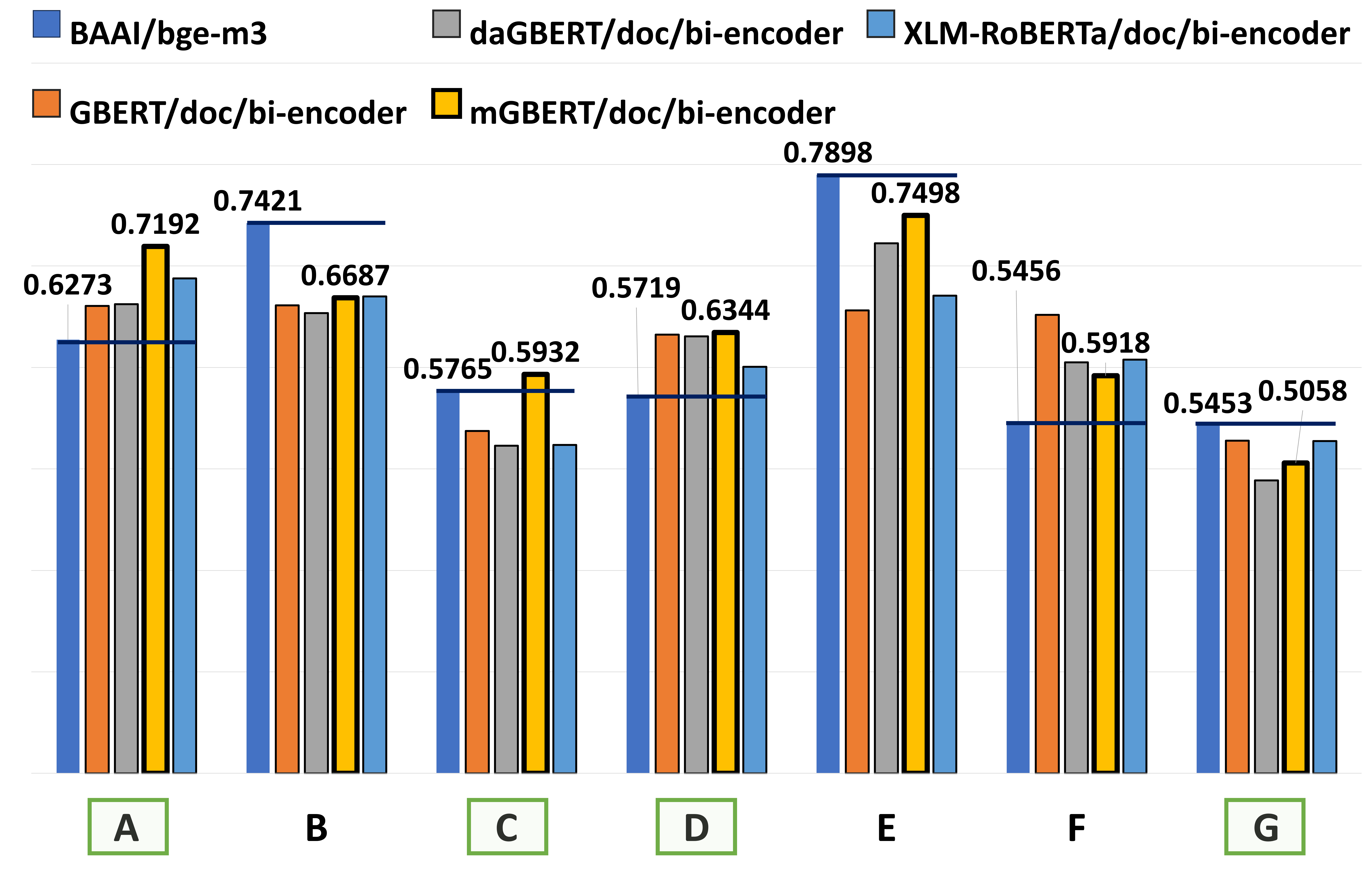}
    \caption{The best-performing fine-tuned mGBERT outperformed the strongest baseline M3 \cite{chen-etal-2024-m3} in almost all plants, the data from which was used for the fine-tuning (i.e., A, C, D, G). }
    \label{fig:results_plant}
\end{figure}

The two-stage fine-tuning process resulted in a systematic performance improvement for three out of four fine-tuned LMs, while the solely bi-encoder fine-tuning improved performance in all LMs compared to the baselines. Our experiments show that using a continually pretrained daGBERT is only beneficial when the domain-specific component is not available in the bi-encoder training data, and the domain-specific knowledge can be compensated for from the domain-pretrained LM. In contrast to the high performance improvement of mBERT, our two-stage fine-tuning approach did not yield significant improvements for GBERT and XLM-RoBERTa. Further experiments with dataset configuration, i.e., composition and size, as well as fine-tuning parameters, are required to investigate the potential performance improvement. 
 
When breaking down the overall results at the plant level, we see that mBERT outperforms our strongest baseline, M3, in three out of four plants, for which the data was used for triplet generation (\Cref{fig:results_plant}). Plant G has the largest number of graph edges in KG after linking prediction; therefore, potentially more triplets should have been used in each of the fine-tuning stages to see a positive effect on this plant from the semantic relations encoded in these connections. 
The fine-tuned models systematically outperform the four plants from the datasets, but further investigation is necessary to refine the methodology and achieve generalizability and optimal performance for each plant.

\subsection{Discussion}
The results showed that leveraging a heterogeneous domain KG that combines signals from text logs and FLs significantly enhances document representations for the process industry domain. Our results indicate that we have successfully adapted and modified the SciNCL methodology from a domain of scientific publications to the process industry domain. The largest impact was observed in using the document triplets collected using GE to create positive and negative query-document pairs for the bi-encoder fine-tuning. These findings suggest that SciNCL could be further expanded into more heterogeneous graphs with limited text data availability. Furthermore, we demonstrate that triplets obtained through kNN in the GE vector space can be used indirectly as a source of synthetic data for downstream tasks. The current experiments focused solely on creating synthetic data to train a bi-encoder; however, the methodology can be further explored to generate data for auxiliary tasks, such as question answering, document ranking, and text classification.

In future work, we plan to expand our KG by incorporating named entity recognition (NER), which will enable us to enrich the graph with additional entities and relationship types, such as chemicals and products. We also aim to optimize the methodology by initializing the GE model with high-performing text encoders designed for semantic text similarity tasks, such as M3 \cite{chen-etal-2024-m3}. Additionally, we will evaluate the impact of various parameters, including the number of positive and negative documents retrieved using GEs, and explore methods to ensure the quality of the generated triplets. We also plan to assess the effect of using data from other plants as triplet sources,  optimize model training parameters, and explore other strategies for the negative selection in the query-document triplets. Finally, we will focus on improving the domain dataset for bi-encoder fine-tuning by enhancing the domain-specificity and quality of the query-document pairs derived from both publicly available and proprietary sources. Our further analysis will include a detailed investigation of how the methodology impacts performance at the plant level, with the goal of optimizing results for each individual plant.

\section{Conclusion}
This work explores the application of the SciNCL methodology for domain adaptation of LMs in the process industry, focusing on fine-tuning them using triplets derived from the vector representation of the domain knowledge graph. Our experiments show that the query-document triplets generated from the GE play a crucial role in providing domain-specific knowledge during the fine-tuning of the four LMs we evaluated.

\section*{Limitations}
The limitations of this study primarily stem from the restricted scope of the data and methodology. Our experiments were conducted using graph embedding models built from data from only four plants and evaluated on just seven plants. The limited dataset size may affect the generalizability of the results, and the performance observed here may not necessarily hold for larger, more diverse datasets with a different composition of data from plants in the process industry. Variations of training datasets for both document-level fine-tuning and as a bi-encoder can additionally impact the results.

Furthermore, we did not conduct exhaustive parameter tuning of the graph embedding model, triplet selection, or a detailed investigation of the parameters adjusted to fine-tune specific models, which could potentially impact the performance and robustness of the results.  Further investigation is needed to assess the scalability and generalizability of the approach across various conditions, including the connectivity of knowledge graphs, the quality of text logs, the methods used in the preprocessing step for link prediction, and the proportion of subdomains used in training and evaluation (e.g., pharmaceuticals and chemistry).

Additionally, the adaptation of the SciNCL methodology may not produce the same positive trends or results if applied to other industries, domains, or languages. Hence, the approach requires extended investigation when applied to a specific linguistic and domain context. Moreover, our results may not be fully reproducible due to the use of proprietary data, which limits external validation. 

Finally, the methodology described in this paper was not compared to other methods of utilizing KGs for language model fine-tuning or enhancement. A comprehensive comparison was beyond the scope of this research; therefore, we do not rule out the possibility that other existing methods using graph embeddings may outperform SciNCL.

\section*{Acknowledgments}
This project is supported by the Federal Ministry for Economic Affairs and Climate Action (BMWK) on the basis of a decision by the German Bundestag. Additional funding was provided by the Federal Ministry of Education and Research (BMBF) in the form of a fellowship under the IFI program of the German Academic Exchange Service (DAAD). We would like to thank Marco Kaiser and Thomas Walton for their contributions to the link prediction services. We also appreciate Malte Ostendorff for the insightful discussions on solutions towards domain adaptation using SciNCL.

\bibliography{custom}

\begin{thebibliography}{57}
\providecommand{\natexlab}[1]{#1}

\bibitem[{Akhbardeh et~al.(2020)Akhbardeh, Desell, and Zampieri}]{akhbardeh_maintnet_2020}
Farhad Akhbardeh, Travis Desell, and Marcos Zampieri. 2020.
\newblock \href {https://doi.org/10.18653/v1/2020.coling-demos.2} {{MaintNet}: {A} {Collaborative} {Open}-{Source} {Library} for {Predictive} {Maintenance} {Language} {Resources}}.
\newblock In \emph{Proceedings of the 28th {International} {Conference} on {Computational} {Linguistics}: {System} {Demonstrations}}, pages 7--11, Barcelona, Spain (Online). International Committee on Computational Linguistics (ICCL).

\bibitem[{Asada et~al.(2021)Asada, Gunasekaran, Miwa, and Sasaki}]{asada_representing_2021}
Masaki Asada, Nallappan Gunasekaran, Makoto Miwa, and Yutaka Sasaki. 2021.
\newblock \href {https://doi.org/10.3389/frma.2021.670206} {Representing a {Heterogeneous} {Pharmaceutical} {Knowledge}-{Graph} with {Textual} {Information}}.
\newblock \emph{Frontiers in Research Metrics and Analytics}, 6:670206.

\bibitem[{Beltagy et~al.(2019)Beltagy, Lo, and Cohan}]{beltagy_scibert_2019}
Iz~Beltagy, Kyle Lo, and Arman Cohan. 2019.
\newblock \href {https://doi.org/10.18653/v1/D19-1371} {{SciBERT}: {A} {Pretrained} {Language} {Model} for {Scientific} {Text}}.
\newblock In \emph{Proceedings of the 2019 {Conference} on {Empirical} {Methods} in {Natural} {Language} {Processing} and the 9th {International} {Joint} {Conference} on {Natural} {Language} {Processing} ({EMNLP}-{IJCNLP})}, pages 3615--3620, Hong Kong, China. Association for Computational Linguistics.

\bibitem[{Brundage et~al.(2021)Brundage, Sexton, Hodkiewicz, Dima, and Lukens}]{brundage_technical_2021}
Michael~P. Brundage, Thurston Sexton, Melinda Hodkiewicz, Alden Dima, and Sarah Lukens. 2021.
\newblock \href {https://doi.org/10.1016/j.mfglet.2020.11.001} {Technical language processing: {Unlocking} maintenance knowledge}.
\newblock \emph{Manufacturing Letters}, 27:42--46.

\bibitem[{Bucher et~al.(2016)Bucher, Herbin, and Jurie}]{bucher_hard_2016}
Maxime Bucher, Stéphane Herbin, and Frédéric Jurie. 2016.
\newblock \href {https://hal.science/hal-01356757} {Hard {Negative} {Mining} for {Metric} {Learning} {Based} {Zero}-{Shot} {Classification}}.
\newblock In \emph{{ECCV} 16 {WS} {TASK}-{CV}: {Transferring} and {Adapting} {Source} {Knowledge} in {Computer} {Vision}}, Amsterdam, Netherlands.

\bibitem[{Chan et~al.(2020)Chan, Schweter, and Möller}]{chan_germans_2020}
Branden Chan, Stefan Schweter, and Timo Möller. 2020.
\newblock \href {https://doi.org/10.18653/v1/2020.coling-main.598} {German's {Next} {Language} {Model}}.
\newblock In \emph{Proceedings of the 28th {International} {Conference} on {Computational} {Linguistics}}, pages 6788--6796, Barcelona, Spain (Online). International Committee on Computational Linguistics.

\bibitem[{Chen et~al.(2024)Chen, Xiao, Zhang, Luo, Lian, and Liu}]{chen-etal-2024-m3}
Jianlyu Chen, Shitao Xiao, Peitian Zhang, Kun Luo, Defu Lian, and Zheng Liu. 2024.
\newblock \href {https://doi.org/10.18653/v1/2024.findings-acl.137} {{M}3-embedding: Multi-linguality, multi-functionality, multi-granularity text embeddings through self-knowledge distillation}.
\newblock In \emph{Findings of the Association for Computational Linguistics: ACL 2024}, pages 2318--2335, Bangkok, Thailand. Association for Computational Linguistics.

\bibitem[{Cohan et~al.(2020)Cohan, Feldman, Beltagy, Downey, and Weld}]{cohan_specter_2020}
Arman Cohan, Sergey Feldman, Iz~Beltagy, Doug Downey, and Daniel Weld. 2020.
\newblock \href {https://doi.org/10.18653/v1/2020.acl-main.207} {{SPECTER}: {Document}-level {Representation} {Learning} using {Citation}-informed {Transformers}}.
\newblock In \emph{Proceedings of the 58th {Annual} {Meeting} of the {Association} for {Computational} {Linguistics}}, pages 2270--2282, Online. Association for Computational Linguistics.

\bibitem[{Conneau et~al.(2020)Conneau, Khandelwal, Goyal, Chaudhary, Wenzek, Guzm{\'a}n, Grave, Ott, Zettlemoyer, and Stoyanov}]{conneau-etal-2020-unsupervised}
Alexis Conneau, Kartikay Khandelwal, Naman Goyal, Vishrav Chaudhary, Guillaume Wenzek, Francisco Guzm{\'a}n, Edouard Grave, Myle Ott, Luke Zettlemoyer, and Veselin Stoyanov. 2020.
\newblock \href {https://doi.org/10.18653/v1/2020.acl-main.747} {Unsupervised cross-lingual representation learning at scale}.
\newblock In \emph{Proceedings of the 58th Annual Meeting of the Association for Computational Linguistics}, pages 8440--8451, Online. Association for Computational Linguistics.

\bibitem[{Devlin et~al.(2019)Devlin, Chang, Lee, and Toutanova}]{devlin-etal-2019-bert}
Jacob Devlin, Ming-Wei Chang, Kenton Lee, and Kristina Toutanova. 2019.
\newblock \href {https://doi.org/10.18653/v1/N19-1423} {{BERT}: Pre-training of deep bidirectional transformers for language understanding}.
\newblock In \emph{Proceedings of the 2019 Conference of the North {A}merican Chapter of the Association for Computational Linguistics: Human Language Technologies, Volume 1 (Long and Short Papers)}, pages 4171--4186, Minneapolis, Minnesota. Association for Computational Linguistics.

\bibitem[{Dima et~al.(2021)Dima, Lukens, Hodkiewicz, Sexton, and Brundage}]{dima_adapting_2021}
Alden Dima, Sarah Lukens, Melinda Hodkiewicz, Thurston Sexton, and Michael~P. Brundage. 2021.
\newblock \href {https://doi.org/10.1002/ail2.33} {Adapting natural language processing for technical text}.
\newblock \emph{Applied AI Letters}, 2(3):e33.
\newblock \_eprint: https://onlinelibrary.wiley.com/doi/pdf/10.1002/ail2.33.

\bibitem[{Granite Embedding~Team(2024)}]{granite2024embedding}
IBM Granite Embedding~Team. 2024.
\newblock \href {https://github.com/ibm-granite/granite-embedding-models/} {Granite embedding models}.

\bibitem[{Guo and Yu(2022)}]{guo_domain_2022}
Xu~Guo and Han Yu. 2022.
\newblock \href {https://doi.org/10.48550/arXiv.2211.03154} {On the {Domain} {Adaptation} and {Generalization} of {Pretrained} {Language} {Models}: {A} {Survey}}.
\newblock \emph{arXiv preprint}.
\newblock ArXiv:2211.03154 [cs].

\bibitem[{Gururangan et~al.(2020)Gururangan, Marasović, Swayamdipta, Lo, Beltagy, Downey, and Smith}]{gururangan_dont_2020}
Suchin Gururangan, Ana Marasović, Swabha Swayamdipta, Kyle Lo, Iz~Beltagy, Doug Downey, and Noah~A. Smith. 2020.
\newblock \href {https://doi.org/10.18653/v1/2020.acl-main.740} {Don't {Stop} {Pretraining}: {Adapt} {Language} {Models} to {Domains} and {Tasks}}.
\newblock In \emph{Proceedings of the 58th {Annual} {Meeting} of the {Association} for {Computational} {Linguistics}}, pages 8342--8360, Online. Association for Computational Linguistics.

\bibitem[{Huang et~al.(2019)Huang, Altosaar, and Ranganath}]{huang_clinicalbert_2019}
Kexin Huang, Jaan Altosaar, and Rajesh Ranganath. 2019.
\newblock \href {http://arxiv.org/abs/1904.05342} {{ClinicalBERT}: {Modeling} {Clinical} {Notes} and {Predicting} {Hospital} {Readmission}}.
\newblock \emph{CoRR}, abs/1904.05342.
\newblock ArXiv: 1904.05342.

\bibitem[{Johnson et~al.(2021)Johnson, Douze, and Jégou}]{johnson_billion-scale_2021}
Jeff Johnson, Matthijs Douze, and Hervé Jégou. 2021.
\newblock \href {https://doi.org/10.1109/TBDATA.2019.2921572} {Billion-{Scale} {Similarity} {Search} with {GPUs}}.
\newblock \emph{IEEE Transactions on Big Data}, 7(3):535--547.
\newblock Conference Name: IEEE Transactions on Big Data.

\bibitem[{Kinney et~al.(2023)Kinney, Anastasiades, Authur, Beltagy, Bragg, Buraczynski, Cachola, Candra, Chandrasekhar, Cohan, Crawford, Downey, Dunkelberger, Etzioni, Evans, Feldman, Gorney, Graham, Hu, Huff, King, Kohlmeier, Kuehl, Langan, Lin, Liu, Lo, Lochner, MacMillan, Murray, Newell, Rao, Rohatgi, Sayre, Shen, Singh, Soldaini, Subramanian, Tanaka, Wade, Wagner, Wang, Wilhelm, Wu, Yang, Zamarron, Van~Zuylen, and Weld}]{kinney_semantic_2023}
Rodney Kinney, Chloe Anastasiades, Russell Authur, Iz~Beltagy, Jonathan Bragg, Alexandra Buraczynski, Isabel Cachola, Stefan Candra, Yoganand Chandrasekhar, Arman Cohan, Miles Crawford, Doug Downey, Jason Dunkelberger, Oren Etzioni, Rob Evans, Sergey Feldman, Joseph Gorney, David Graham, Fangzhou Hu, and 29 others. 2023.
\newblock \href {http://arxiv.org/abs/2301.10140} {The {Semantic} {Scholar} {Open} {Data} {Platform}}.
\newblock \emph{arXiv preprint}.
\newblock ArXiv:2301.10140 [cs].

\bibitem[{Lee et~al.(2019)Lee, Yoon, Kim, Kim, Kim, So, and Kang}]{lee_biobert_2019}
Jinhyuk Lee, Wonjin Yoon, Sungdong Kim, Donghyeon Kim, Sunkyu Kim, Chan~Ho So, and Jaewoo Kang. 2019.
\newblock \href {https://doi.org/10.1093/bioinformatics/btz682} {{BioBERT}: a pre-trained biomedical language representation model for biomedical text mining}.
\newblock \emph{Bioinformatics}, 36(4):1234--1240.

\bibitem[{Lee et~al.(2024)Lee, Shakir, Koenig, and Lipp}]{germanemb2024mxbai}
Sean Lee, Aamir Shakir, Darius Koenig, and Julius Lipp. 2024.
\newblock \href {https://www.mixedbread.ai/blog/deepset-mxbai-embed-de-large-v1} {Open source gets de-licious: Mixedbread x deepset german/english embeddings}.

\bibitem[{Lerer et~al.(2019)Lerer, Wu, Shen, Lacroix, Wehrstedt, Bose, and Peysakhovich}]{lerer_pytorch-biggraph_2019}
Adam Lerer, Ledell Wu, Jiajun Shen, Timothee Lacroix, Luca Wehrstedt, Abhijit Bose, and Alex Peysakhovich. 2019.
\newblock \href {https://proceedings.mlsys.org/paper_files/paper/2019/hash/1eb34d662b67a14e3511d0dfd78669be-Abstract.html} {Pytorch-{BigGraph}: {A} {Large} {Scale} {Graph} {Embedding} {System}}.
\newblock In \emph{Proceedings of the 2nd {Conference} on {Systems} and {Machine} {Learning}}, volume~1, pages 120--131, Palo Alto, CA, USA. mlsys.org.

\bibitem[{Liu et~al.(2020)Liu, Zhou, Zhao, Wang, Ju, Deng, and Wang}]{liu_k-bert_2020}
Weijie Liu, Peng Zhou, Zhe Zhao, Zhiruo Wang, Qi~Ju, Haotang Deng, and Ping Wang. 2020.
\newblock \href {https://doi.org/10.1609/aaai.v34i03.5681} {K-{BERT}: {Enabling} {Language} {Representation} with {Knowledge} {Graph}}.
\newblock \emph{Proceedings of the AAAI Conference on Artificial Intelligence}, 34(03):2901--2908.
\newblock Number: 03.

\bibitem[{Liu et~al.(2019)Liu, Ott, Goyal, Du, Joshi, Chen, Levy, Lewis, Zettlemoyer, and Stoyanov}]{liu_roberta_2019}
Yinhan Liu, Myle Ott, Naman Goyal, Jingfei Du, Mandar Joshi, Danqi Chen, Omer Levy, Mike Lewis, Luke Zettlemoyer, and Veselin Stoyanov. 2019.
\newblock \href {https://doi.org/10.48550/arXiv.1907.11692} {{RoBERTa}: {A} {Robustly} {Optimized} {BERT} {Pretraining} {Approach}}.
\newblock \emph{arXiv preprint}.
\newblock ArXiv:1907.11692 [cs].

\bibitem[{Lo et~al.(2020)Lo, Wang, Neumann, Kinney, and Weld}]{lo_s2orc_2020}
Kyle Lo, Lucy~Lu Wang, Mark Neumann, Rodney Kinney, and Daniel Weld. 2020.
\newblock \href {https://doi.org/10.18653/v1/2020.acl-main.447} {{S2ORC}: {The} {Semantic} {Scholar} {Open} {Research} {Corpus}}.
\newblock In \emph{Proceedings of the 58th {Annual} {Meeting} of the {Association} for {Computational} {Linguistics}}, pages 4969--4983, Online. Association for Computational Linguistics.

\bibitem[{May et~al.(2022)May, Neidhöfer, Körner, Schäfer, and Lanza}]{may_applying_2022}
Marvin~Carl May, Jan Neidhöfer, Tom Körner, Louis Schäfer, and Gisela Lanza. 2022.
\newblock \href {https://doi.org/10.1016/j.procir.2022.10.071} {Applying {Natural} {Language} {Processing} in {Manufacturing}}.
\newblock \emph{Procedia CIRP}, 115:184--189.

\bibitem[{Naqvi et~al.(2024)Naqvi, Ghufran, Varnier, Nicod, Javed, and Zerhouni}]{naqvi_unlocking_2024}
Syed Meesam~Raza Naqvi, Mohammad Ghufran, Christophe Varnier, Jean-Marc Nicod, Kamran Javed, and Noureddine Zerhouni. 2024.
\newblock \href {https://doi.org/10.1016/j.compind.2024.104083} {Unlocking maintenance insights in industrial text through semantic search}.
\newblock \emph{Computers in Industry}, 157-158:104083.

\bibitem[{Naqvi et~al.(2022)Naqvi, Varnier, Nicod, Zerhouni, and Ghufran}]{naqvi_leveraging_2022}
Syed Meesam~Raza Naqvi, Christophe Varnier, Jean-Marc Nicod, Noureddine Zerhouni, and Mohammad Ghufran. 2022.
\newblock \href {https://doi.org/10.1007/978-3-030-98531-8_7} {Leveraging {Free}-{Form} {Text} in {Maintenance} {Logs} {Through} {BERT} {Transfer} {Learning}}.
\newblock In \emph{Progresses in {Artificial} {Intelligence} \& {Robotics}: {Algorithms} \& {Applications}}, pages 63--75, Cham. Springer International Publishing.

\bibitem[{Nguyen et~al.(2016)Nguyen, Rosenberg, Song, Gao, Tiwary, Majumder, and Deng}]{NguyenRSGTMD16}
Tri Nguyen, Mir Rosenberg, Xia Song, Jianfeng Gao, Saurabh Tiwary, Rangan Majumder, and Li~Deng. 2016.
\newblock \href {https://ceur-ws.org/Vol-1773/CoCoNIPS\_2016\_paper9.pdf} {{MS} {MARCO:} {A} human generated machine reading comprehension dataset}.
\newblock In \emph{Proceedings of the Workshop on Cognitive Computation: Integrating neural and symbolic approaches 2016 co-located with the 30th Annual Conference on Neural Information Processing Systems {(NIPS} 2016), Barcelona, Spain, December 9, 2016}, volume 1773 of \emph{{CEUR} Workshop Proceedings}. CEUR-WS.org.

\bibitem[{Nussbaum and Duderstadt(2025)}]{nussbaum2025trainingsparsemixtureexperts}
Zach Nussbaum and Brandon Duderstadt. 2025.
\newblock \href {https://arxiv.org/abs/2502.07972} {Training sparse mixture of experts text embedding models}.
\newblock \emph{Preprint}, arXiv:2502.07972.

\bibitem[{OpenAI(2024)}]{openai2024embedding}
OpenAI. 2024.
\newblock \href {https://openai.com/index/new-embedding-models-and-api-updates/} {New embedding models and api updates}.

\bibitem[{Ostendorff et~al.(2022)Ostendorff, Rethmeier, Augenstein, Gipp, and Rehm}]{ostendorff_neighborhood_2022}
Malte Ostendorff, Nils Rethmeier, Isabelle Augenstein, Bela Gipp, and Georg Rehm. 2022.
\newblock \href {https://doi.org/10.18653/v1/2022.emnlp-main.802} {Neighborhood {Contrastive} {Learning} for {Scientific} {Document} {Representations} with {Citation} {Embeddings}}.
\newblock In \emph{Proceedings of the 2022 {Conference} on {Empirical} {Methods} in {Natural} {Language} {Processing}}, pages 11670--11688, Abu Dhabi, United Arab Emirates. Association for Computational Linguistics.

\bibitem[{Peters et~al.(2019)Peters, Neumann, Logan, Schwartz, Joshi, Singh, and Smith}]{peters_knowledge_2019}
Matthew~E. Peters, Mark Neumann, Robert Logan, Roy Schwartz, Vidur Joshi, Sameer Singh, and Noah~A. Smith. 2019.
\newblock \href {https://doi.org/10.18653/v1/D19-1005} {Knowledge {Enhanced} {Contextual} {Word} {Representations}}.
\newblock In \emph{Proceedings of the 2019 {Conference} on {Empirical} {Methods} in {Natural} {Language} {Processing} and the 9th {International} {Joint} {Conference} on {Natural} {Language} {Processing} ({EMNLP}-{IJCNLP})}, pages 43--54, Hong Kong, China. Association for Computational Linguistics.

\bibitem[{Qiu et~al.(2020)Qiu, Sun, Xu, Shao, Dai, and Huang}]{qiu_pre-trained_2020}
XiPeng Qiu, TianXiang Sun, YiGe Xu, YunFan Shao, Ning Dai, and XuanJing Huang. 2020.
\newblock \href {https://doi.org/10.1007/s11431-020-1647-3} {Pre-trained models for natural language processing: {A} survey}.
\newblock \emph{Science China Technological Sciences}, 63(10):1872--1897.

\bibitem[{Reimers and Gurevych(2019)}]{reimers_sentence-bert_2019}
Nils Reimers and Iryna Gurevych. 2019.
\newblock \href {https://doi.org/10.18653/v1/D19-1410} {Sentence-{BERT}: {Sentence} {Embeddings} using {Siamese} {BERT}-{Networks}}.
\newblock In \emph{Proceedings of the 2019 {Conference} on {Empirical} {Methods} in {Natural} {Language} {Processing} and the 9th {International} {Joint} {Conference} on {Natural} {Language} {Processing} ({EMNLP}-{IJCNLP})}, pages 3982--3992, Hong Kong, China. Association for Computational Linguistics.

\bibitem[{Saunshi et~al.(2019)Saunshi, Plevrakis, Arora, Khodak, and Khandeparkar}]{saunshi_theoretical_2019}
Nikunj Saunshi, Orestis Plevrakis, Sanjeev Arora, Mikhail Khodak, and Hrishikesh Khandeparkar. 2019.
\newblock \href {https://proceedings.mlr.press/v97/saunshi19a.html} {A {Theoretical} {Analysis} of {Contrastive} {Unsupervised} {Representation} {Learning}}.
\newblock In \emph{Proceedings of the 36th {International} {Conference} on {Machine} {Learning}}, pages 5628--5637. PMLR.
\newblock ISSN: 2640-3498.

\bibitem[{Schroff et~al.(2015)Schroff, Kalenichenko, and Philbin}]{schroff_facenet_2015}
Florian Schroff, Dmitry Kalenichenko, and James Philbin. 2015.
\newblock \href {https://doi.org/10.1109/CVPR.2015.7298682} {{FaceNet}: {A} unified embedding for face recognition and clustering}.
\newblock In \emph{2015 {IEEE} {Conference} on {Computer} {Vision} and {Pattern} {Recognition} ({CVPR})}, pages 815--823.
\newblock ISSN: 1063-6919.

\bibitem[{Sentence-Transformers(2021)}]{sentence_transformers_msmarco_hard_negatives}
Sentence-Transformers. 2021.
\newblock Ms marco passage hard negatives.
\newblock \url{https://huggingface.co/datasets/sentence-transformers/msmarco-hard-negatives}.

\bibitem[{Stewart et~al.(2022)Stewart, Hodkiewicz, Liu, and French}]{stewart_mwo2kg_2022}
Michael Stewart, Melinda Hodkiewicz, Wei Liu, and Tim French. 2022.
\newblock \href {https://doi.org/10.1177/1748006X221131128} {{MWO2KG} and {Echidna}: {Constructing} and exploring knowledge graphs from maintenance data}.
\newblock \emph{Proceedings of the Institution of Mechanical Engineers, Part O: Journal of Risk and Reliability}, page 1748006X221131128.
\newblock Publisher: SAGE Publications.

\bibitem[{Strubell et~al.(2020)Strubell, Ganesh, and McCallum}]{strubell_energy_2020}
Emma Strubell, Ananya Ganesh, and Andrew McCallum. 2020.
\newblock \href {https://doi.org/10.1609/aaai.v34i09.7123} {Energy and {Policy} {Considerations} for {Modern} {Deep} {Learning} {Research}}.
\newblock \emph{Proceedings of the AAAI Conference on Artificial Intelligence}, 34(09):13693--13696.
\newblock Number: 09.

\bibitem[{Thakur et~al.(2021)Thakur, Reimers, R\"{u}ckl\'{e}, Srivastava, and Gurevych}]{beir_2021}
Nandan Thakur, Nils Reimers, Andreas R\"{u}ckl\'{e}, Abhishek Srivastava, and Iryna Gurevych. 2021.
\newblock \href {https://datasets-benchmarks-proceedings.neurips.cc/paper_files/paper/2021/file/65b9eea6e1cc6bb9f0cd2a47751a186f-Paper-round2.pdf} {Beir: A heterogeneous benchmark for zero-shot evaluation of information retrieval models}.
\newblock In \emph{Proceedings of the Neural Information Processing Systems Track on Datasets and Benchmarks}, volume~1.

\bibitem[{Usuga~Cadavid et~al.(2020)Usuga~Cadavid, Grabot, Lamouri, Pellerin, and Fortin}]{usuga_cadavid_valuing_2020}
Juan~Pablo Usuga~Cadavid, Bernard Grabot, Samir Lamouri, Robert Pellerin, and Arnaud Fortin. 2020.
\newblock \href {https://doi.org/10.1080/17517575.2020.1790043} {Valuing free-form text data from maintenance logs through transfer learning with {CamemBERT}}.
\newblock \emph{Enterprise Information Systems}, 16(6):1790043.
\newblock Publisher: Taylor \& Francis \_eprint: https://doi.org/10.1080/17517575.2020.1790043.

\bibitem[{Usuga-Cadavid et~al.(2022)Usuga-Cadavid, Lamouri, Grabot, and Fortin}]{usuga-cadavid_using_2022}
Juan~Pablo Usuga-Cadavid, Samir Lamouri, Bernard Grabot, and Arnaud Fortin. 2022.
\newblock \href {https://doi.org/10.1080/00207543.2021.1951868} {Using deep learning to value free-form text data for predictive maintenance}.
\newblock \emph{International Journal of Production Research}, 60(14):4548--4575.

\bibitem[{Wang et~al.(2024)Wang, Yang, Huang, Yang, Majumder, and Wei}]{wang2024multilinguale5textembeddings}
Liang Wang, Nan Yang, Xiaolong Huang, Linjun Yang, Rangan Majumder, and Furu Wei. 2024.
\newblock \href {https://arxiv.org/abs/2402.05672} {Multilingual e5 text embeddings: A technical report}.
\newblock \emph{Preprint}, arXiv:2402.05672.

\bibitem[{Wang and Isola(2020)}]{wang_understanding_2020}
Tongzhou Wang and Phillip Isola. 2020.
\newblock \href {https://proceedings.mlr.press/v119/wang20k.html} {Understanding {Contrastive} {Representation} {Learning} through {Alignment} and {Uniformity} on the {Hypersphere}}.
\newblock In \emph{Proceedings of the 37th {International} {Conference} on {Machine} {Learning}}, pages 9929--9939. PMLR.
\newblock ISSN: 2640-3498.

\bibitem[{Wang et~al.(2021)Wang, Gao, Zhu, Zhang, Liu, Li, and Tang}]{wang_kepler_2021}
Xiaozhi Wang, Tianyu Gao, Zhaocheng Zhu, Zhengyan Zhang, Zhiyuan Liu, Juanzi Li, and Jian Tang. 2021.
\newblock \href {https://doi.org/10.1162/tacl_a_00360} {{KEPLER}: {A} {Unified} {Model} for {Knowledge} {Embedding} and {Pre}-trained {Language} {Representation}}.
\newblock \emph{Transactions of the Association for Computational Linguistics}, 9:176--194.
\newblock Place: Cambridge, MA Publisher: MIT Press.

\bibitem[{Wu et~al.(2017)Wu, Manmatha, Smola, and Krahenbuhl}]{wu_sampling_2017}
Chao-Yuan Wu, R.~Manmatha, Alexander~J. Smola, and Philipp Krahenbuhl. 2017.
\newblock \href {https://doi.org/10.1109/ICCV.2017.309} {Sampling {Matters} in {Deep} {Embedding} {Learning}}.
\newblock In \emph{2017 {IEEE} {International} {Conference} on {Computer} {Vision} ({ICCV})}, pages 2859--2867, Venice. IEEE.

\bibitem[{Xia et~al.(2023)Xia, Liang, Leng, and Zheng}]{xia_maintenance_2023}
Liqiao Xia, Yongshi Liang, Jiewu Leng, and Pai Zheng. 2023.
\newblock \href {https://doi.org/10.1016/j.ress.2022.109068} {Maintenance planning recommendation of complex industrial equipment based on knowledge graph and graph neural network}.
\newblock \emph{Reliability Engineering \& System Safety}, 232:109068.

\bibitem[{Xiao et~al.(2023)Xiao, Zheng, Shi, Du, and Hong}]{xiao_knowledge_2023}
Youzi Xiao, Shuai Zheng, Jiancheng Shi, Xiaodong Du, and Jun Hong. 2023.
\newblock \href {https://doi.org/10.1016/j.jmsy.2023.08.006} {Knowledge graph-based manufacturing process planning: {A} state-of-the-art review}.
\newblock \emph{Journal of Manufacturing Systems}, 70:417--435.

\bibitem[{Yasunaga et~al.(2022)Yasunaga, Leskovec, and Liang}]{yasunaga_linkbert_2022}
Michihiro Yasunaga, Jure Leskovec, and Percy Liang. 2022.
\newblock \href {https://doi.org/10.18653/v1/2022.acl-long.551} {{LinkBERT}: {Pretraining} {Language} {Models} with {Document} {Links}}.
\newblock In \emph{Proceedings of the 60th {Annual} {Meeting} of the {Association} for {Computational} {Linguistics} ({Volume} 1: {Long} {Papers})}, pages 8003--8016, Dublin, Ireland. Association for Computational Linguistics.

\bibitem[{Zhang et~al.(2024)Zhang, Zhang, Long, Xie, Dai, Tang, Lin, Yang, Xie, Huang, Zhang, Li, and Zhang}]{zhang-etal-2024-mgte}
Xin Zhang, Yanzhao Zhang, Dingkun Long, Wen Xie, Ziqi Dai, Jialong Tang, Huan Lin, Baosong Yang, Pengjun Xie, Fei Huang, Meishan Zhang, Wenjie Li, and Min Zhang. 2024.
\newblock \href {https://doi.org/10.18653/v1/2024.emnlp-industry.103} {{mGTE}: Generalized long-context text representation and reranking models for multilingual text retrieval}.
\newblock In \emph{Proceedings of the 2024 Conference on Empirical Methods in Natural Language Processing: Industry Track}, pages 1393--1412, Miami, Florida, US. Association for Computational Linguistics.

\bibitem[{Zhang et~al.(2023)Zhang, Malkov, Florez, Park, McWilliams, Han, and El-Kishky}]{zhang_twhin-bert_2023}
Xinyang Zhang, Yury Malkov, Omar Florez, Serim Park, Brian McWilliams, Jiawei Han, and Ahmed El-Kishky. 2023.
\newblock \href {https://doi.org/10.1145/3580305.3599921} {{TwHIN}-{BERT}: {A} {Socially}-{Enriched} {Pre}-trained {Language} {Model} for {Multilingual} {Tweet} {Representations} at {Twitter}}.
\newblock In \emph{Proceedings of the 29th {ACM} {SIGKDD} {Conference} on {Knowledge} {Discovery} and {Data} {Mining}}, pages 5597--5607, Long Beach CA USA. ACM.

\bibitem[{Zhang et~al.(2022)Zhang, Shen, Wu, Xie, Hao, Wang, Wang, and Han}]{zhang_metadata-induced_2022}
Yu~Zhang, Zhihong Shen, Chieh-Han Wu, Boya Xie, Junheng Hao, Ye-Yi Wang, Kuansan Wang, and Jiawei Han. 2022.
\newblock \href {https://doi.org/10.1145/3485447.3512174} {Metadata-{Induced} {Contrastive} {Learning} for {Zero}-{Shot} {Multi}-{Label} {Text} {Classification}}.
\newblock In \emph{Proceedings of the {ACM} {Web} {Conference} 2022}, {WWW} '22, pages 3162--3173, New York, NY, USA. Association for Computing Machinery.

\bibitem[{Zhang et~al.(2019)Zhang, Han, Liu, Jiang, Sun, and Liu}]{zhang_ernie_2019}
Zhengyan Zhang, Xu~Han, Zhiyuan Liu, Xin Jiang, Maosong Sun, and Qun Liu. 2019.
\newblock \href {https://doi.org/10.18653/v1/P19-1139} {{ERNIE}: {Enhanced} {Language} {Representation} with {Informative} {Entities}}.
\newblock In \emph{Proceedings of the 57th {Annual} {Meeting} of the {Association} for {Computational} {Linguistics}}, pages 1441--1451, Florence, Italy. Association for Computational Linguistics.

\bibitem[{Zhong et~al.(2024)Zhong, Jackson, West, and Cosma}]{zhong_natural_2024}
Keyi Zhong, Tom Jackson, Andrew West, and Georgina Cosma. 2024.
\newblock \href {https://doi.org/10.1016/j.procs.2024.02.029} {Natural {Language} {Processing} {Approaches} in {Industrial} {Maintenance}: {A} {Systematic} {Literature} {Review}}.
\newblock \emph{Procedia Computer Science}, 232:2082--2097.

\bibitem[{Zhukova et~al.(2025{\natexlab{a}})Zhukova, Matt, and Gipp}]{zhukova-etal-2025-automated}
Anastasia Zhukova, Christian~E. Matt, and Bela Gipp. 2025{\natexlab{a}}.
\newblock \href {https://aclanthology.org/2025.loreslm-1.8/} {Automated collection of evaluation dataset for semantic search in low-resource domain language}.
\newblock In \emph{Proceedings of the First Workshop on Language Models for Low-Resource Languages}, pages 112--122, Abu Dhabi, United Arab Emirates. Association for Computational Linguistics.

\bibitem[{Zhukova et~al.(2025{\natexlab{b}})Zhukova, Matt, and Gipp}]{Zhukova2025a}
Anastasia Zhukova, Christian~E. Matt, and Bela Gipp. 2025{\natexlab{b}}.
\newblock \href {https://link.springer.com/chapter/10.1007/978-3-032-02551-7\_8} {Efficient domain-adaptive continual pretraining for the process industry in the german language}.
\newblock In \emph{Text, Speech and Dialogue. Proceedings of the 28th International Conference TSD2025, Erlangen, Germany, August 2025}, Cham. Springer Nature Switzerland.

\bibitem[{Zhukova et~al.(2024)Zhukova, von Sperl, Matt, and Gipp}]{Zhukova2024}
Anastasia Zhukova, Lukas von Sperl, Christian~E. Matt, and Bela Gipp. 2024.
\newblock \href {https://doi.org/10.1007/s10115-024-02212-5} {Generative user-experience research for developing domain-specific natural language processing applications}.
\newblock \emph{Knowledge and Information Systems}, 66:7859–7889.

\bibitem[{Zhukova et~al.(2025{\natexlab{c}})Zhukova, Walton, Matt, and Gipp}]{zhukova2025linkpredictioneventlogs}
Anastasia Zhukova, Thomas Walton, Christian~E. Matt, and Bela Gipp. 2025{\natexlab{c}}.
\newblock \href {https://arxiv.org/abs/2508.09096} {Link prediction for event logs in the process industry}.
\newblock \emph{Preprint}, arXiv:2508.09096.

\end{thebibliography}

\appendix

\section{Implementation details}

\subsection{Knowledge graph}
\label{sec:appendex_knowledge}
We build a domain KG by using all FL nodes and \textit{only those text log nodes that contain at least one attached FL}. Although S2ORC used for SciNLC contained titles and abstracts with more than 100 characters \cite{lo_s2orc_2020}, removing all short text logs (<100 chars) from the domain KG for the process industry would filter out almost half of the available documents. \Cref{tab:overview_process_industry_datasets} presents a statistical analysis of the process industry graphs, revealing that the number of direct links between text logs is low (see the \textit{related\_to} edge type) and emphasizing the importance of indirectly connecting text logs through shared FLs. Since every text log in the graph is linked to at least one FL, there is a high number of \textit{reports\_about} edges, significantly improving the overall connectivity of the graph. 

Data analysis revealed that many FLs are referenced in text logs using domain-specific abbreviations (e.g., A11) but are not explicitly linked as attributes (see \Cref{fig:plant_graph}). This lack of direct association limits our implementation's ability to construct a comprehensive domain KG. To address this problem, we implement two domain-specific custom link prediction models to (1) reconstruct the \textit{reports\_about} and \textit{related\_to} connections within the domain KG, (2) expand the context of domain-specific abbreviations. In the context expansion, we replace the abbreviations such as "A11" with a combination of the abbreviation and its description "A11 Pumpe", which was obtained from the linked FL node. As shown in \Cref{tab:overview_process_industry_datasets}, this approach significantly increases the number of graph edges and expands the number of text log nodes, i.e., text logs that previously lacked associated FLs are now integrated into the graph, enhancing its overall structure and connectivity. Moreover, we apply a record linking model \cite{zhukova2025linkpredictioneventlogs} to restore connectivity between the text logs, i.e., when two records reported about a problem and a solution to it, but were not explicitly linked in a database as related. \Cref{tab:overview_process_industry_datasets} shows that the KG after the link prediction includes the largest number of text log nodes from the available raw data.

\begin{table}[]
\scriptsize
    \centering
    \begin{tabular}{llrrrr }
        \hline
        \makecell{KG stage} & Parameters & {A} & {C} & {D} & {G}  \\
        \hline
        \multirow{3}{*}{\makecell[l]{Raw data}} & node candidates &  24K & 229K & 119K & 147K  \\  
          & \hspace{3mm} FLs & 0.5K & 62K & 37K & 35K  \\
        & \hspace{3mm} text logs & 23K & 167K & 82K & 112K  \\
        
        
        \hline
        \multirow{7}{*}{\makecell[l]{Without \\ link pred.}} & nodes & 22K & 140K  & 114K & 125K \\ 
        & \hspace{3mm} FLs & 0.5K & 62K & 37K & 35K  \\
        & \hspace{3mm} text logs & 21.5K & 78K & 77K & 90K  \\
        \cdashline{2-6}  
        & edges & 22K & 152K & 119K & 138K  \\
         & \hspace{3mm} related\_to & 0.1K & 1K & 1K & 0.1K  \\
         & \hspace{3mm} reports\_about & 21.5K & 89K & 81K & 102K  \\
         & \hspace{3mm} part\_of& 0.5K & 62K & 37K & 35K  \\
         \hline
        \multirow{7}{*}{\makecell[l]{After link\\ prediction}} & nodes & 22K & 172K & 117K & 132K \\ 
        & \hspace{3mm} FLs  & 0.5K & 62K & 37K & 35K  \\
        & \hspace{3mm} text logs & 21.5K & 110K & 80K & 96K  \\
        \cdashline{2-6}  
        & edges  & 24.5K & 333K & 246K & 1815K  \\
         & \hspace{3mm} related\_to & 2.5K & 31K & 86K & 1602K  \\ 
         & \hspace{3mm} reports\_about  & 21.5K & 240K & 123K & 178K  \\ 
         & \hspace{3mm} part\_of & 0.5K & 62K & 37K & 35K  \\
        \hline
    \end{tabular}
    \caption{Overview of the plant datasets, which consist of two types of data: text logs of daily operations and a function location tree that describes the hierarchical structure of the plant machinery. To enhance data connectivity, link prediction was performed to reconstruct the \textit{related\_to} and \textit{reports\_about} links. }
    \label{tab:overview_process_industry_datasets}
\end{table}

\subsection{Graph embeddings}
\label{sec_appendix_graph_emb}
Following SciNCL, we employ PyTorch BigGraph (PBG) \cite{lerer_pytorch-biggraph_2019} for training graph embedding models. PBG supports graphs with multiple node and edge types, making it suitable for the heterogeneous process industry graph introduced in this work. 
PBG processes the input graph as a list of edges, where each edge is defined by a source node, a target node, and an edge type. It generates embeddings for each node, ensuring that connected nodes are positioned closer together in the vector space while unconnected nodes are pushed farther apart. This design ensures that entities with similar neighbor distributions are located near one another in the embedding space. In addition to handling heterogeneous graphs, PBG is highly scalable, making it well-suited to accommodate the growing volume of text logs in the future.

\begin{table}[]
\small
    \centering
    \begin{tabular}{ ccrrrr}
        \hline
        \makecell[c]{Init. with\\sent.\\embed. } & \makecell[c]{LP} & {MRR} & {Hits@1} & {Hits@10} & {AUC} \\
        \hline
        no & no & 16.96 & 09.39 & 30.47 & 59.93 \\
        \hline
        yes & no & 36.48 & 21.54 & 72.25 & 78.75  \\        
        \hline
        yes & yes & 47.80 & 33.91 & 78.21 & 85.06 \\
        \hline
    \end{tabular}
    \caption{Evaluation of graph embedding models. Link prediction (LP) performance of PyTorch BigGraph embeddings trained on the process industry graphs for plant D (1) with and without node initialization with sentence embedding, and (2) before and after data enrichment.}
    \label{tab:biggraph_evaluation_performance}
\end{table}

Our methodology primarily adopts the training parameters for the graph embedding model described in SciNCL, while experimenting with variations in the number of training epochs and using sentence embedding initialization. Some FLs have similar descriptions, such as reactors of different types. Initializing the graph network with semantically meaningful vectors can enhance the learning of structural connections between nodes. For example, by initializing FL nodes using their descriptions, similar FL types (e.g., pumps) are naturally positioned closer together in the embedding space. During training, text logs referencing different pumps are more likely to be mapped to similar regions within the embedding space, as their corresponding FLs were initially placed near each other. Consequently, text logs that do not share the same FLs but originate from semantically similar contexts can still achieve similar node embeddings. Moreover, if text logs contain terms similar to those in the descriptions of FLs, such nodes will also be positioned closely in the vector space. We anticipate that integrating domain-specific relationships with semantic similarity will help mitigate challenges such as low graph connectivity and poor text quality, while improving the sampling of both similar and dissimilar text documents.

The node embeddings for the graph model are initialized by vectorizing the text logs and functional location (FL) descriptions using sentence transformer embeddings from {\small\texttt{PM-AI/bi-encoder\_msmarco\_bert-base\_german}}. \Cref{tab:biggraph_evaluation_performance} presents the link prediction performance for plant D, evaluated using metrics such as MRR, Hits@1, Hits@10, and AUC. For these experiments, 99\% of the data was used for training, while the remaining 1\% (1193 edges) was reserved for testing\footnote{Metric definitions are available in the PBG documentation: \url{https://torchbiggraph.readthedocs.io/en/latest/evaluation.html}}. The results demonstrate that initializing nodes with text embeddings significantly enhances the performance of link prediction.  These findings highlight the crucial role of initializing node embeddings with text embeddings in effectively training graph embedding models within the process industry.

\subsection{Triplet sampling}
\label{sec_appendix_triplet}
For each query document, we sampled two triplets, consisting of one easy negative and one hard negative. This provides a diverse similarity signal while ensuring broader document coverage, even within smaller data subsets. 

To identify text logs with a similar graph context — those associated with similar text logs and functional locations — an approximate nearest neighbor (ANN) search was conducted in the text log embedding space. FAISS \cite{johnson_billion-scale_2021} was employed to construct a flat index of text logs, keyed by their node embeddings. To ensure consistency in distance measurement, embeddings were L2-normalized, and cosine similarity was used as the distance metric by computing the inner product of the embeddings. After building and populating the FAISS index with text log embeddings, queries in the form of text log embeddings $\mathbf{d}^Q$ were used to retrieve the $k$-nearest neighbors based on their cosine distance in the embedding space.

Initial experiments revealed that including overly short text sequences during training significantly degrades language model performance, leading to the introduction of a \textit{minimum text length} criterion. To preserve as many nodes and edges as possible for graph model training, this criterion was applied \textit{after training the graph embedding} model but \textit{before triplet sampling}. By filtering text logs shorter than 100 characters (approximately 15–25 words) only before the ANN search, the approach ensures robust text embeddings, enhances contextual information, and improves node embedding quality in graphs with low edge density. \Cref{tab:overview_process_industry_datasets} shows the number of generated triplets per plant. 

\textit{Positive samples} $d^+$ should be semantically similar to the query document $d^Q$, but not overly similar, to avoid gradient collapse. Additionally, positives should be sampled from a comparable distance to the query embedding $\mathbf{d}^Q$ to ensure balanced training \cite{wang_understanding_2020}. To achieve this, we follow SciNCL that selects positive (similar) documents by first defining the distance to the query embedding $\mathbf{d}^Q$ using $k^+$ and then sampling the $c^+$ nearest neighbors within a range ($k^+ - c^+$, $k^+$], as illustrated by the green band in \Cref{fig:proposed_methodology}.

\textit{Negative samples} $d^-$ should be semantically dissimilar from the query document $d^Q$, meaning they are sampled from a distant region relative to the query embedding  $\mathbf{d}^Q$. Sampling hard-to-learn negatives, i.e., those close to potential positives, has been shown to enhance contrastive learning \cite{bucher_hard_2016, wu_sampling_2017}. However, when negative samples overlap with positive ones, it can introduce noise into the learning signal \cite{saunshi_theoretical_2019}. To prevent such collisions, SciNCL employs a sample-induced margin defined by the $k$ parameter. This margin ensures the sampling of hard negatives (represented by the red band in \Cref{fig:proposed_methodology}) without overlapping with the sampled positives (green band). To provide a robust similarity signal for contrastive learning, SciNCL recommends using a mix of hard and easy negatives, where easy negatives are embeddings located further from the query embedding than hard negatives (outside the red band).

Sampling strategies differ between positive and types of negative samples, i.e., similar to SciNCL. We use kNN sampling for positives and hard negatives, and filtered random sampling for easy negatives. \textit{kNN sampling} is conducted as k-nearest neighbors search $kNN(f_{gem}(\mathbf{d}^Q), \mathbf{D})$ given a graph embedding model (e.g., PyTorch BigGraph \cite{lerer_pytorch-biggraph_2019}) denoted as  $f_{gem}$, document node embeddings $\mathbf{D}$, and a search index (e.g., FAISS \cite{johnson_billion-scale_2021}). From the neighbors around the query document $d^Q$, $c'$ samples are selected using the interval $(k - c', k]$, where $N=\{n_1, n_2, n_3, \dots \}$ represents the neighbors, i.e., $n_i$ is the $i$-th nearest neighbor in the graph embedding space. For example, if $c'=3$ and $k=10$, the selected samples would be the three closest neighbors within the specified range: $n_8, n_9$, and $n_{10}$. \textit{Filtered random} sampling utilizes random sampling, i.e., samples $c'$ documents from the corpus without replacement, but excludes the documents retrieved by kNN, i.e., neighbors within the largest $k$.

Due to the low number of direct connections between text logs, we select the closest $k=2$ neighbors as positive samples. A small $k^+$ ensures that text logs that are positioned close in the graph embedding space, due to their direct links, are included. A higher $k^+$ value could risk skipping such direct links. Similarly, $k_{hard}^{-}$ was set to 50 to ensure that the selected hard negatives were sufficiently dissimilar, as the 50th neighbor is expected to fall outside the direct neighborhood of the query log.

\subsection{Domain-related MS MARCO Dataset}
\label{sec:appendix_dataset}
We created a domain-related version of MS MARCO by running the document collection through our binary classifier, which identifies whether a document is domain-related or not. This classifier is a fine-tuned SciBERT model \cite{beltagy_scibert_2019} trained on an 80K dataset that combines \href{https://huggingface.co/datasets/allenai/scirepeval/viewer/fos}{fields of study (FoS)} data and our in-domain data. The train dataset was split so that 50\% consisted of domain-related texts, selected by the labels in FoS that correspond to our domain, and in-domain texts, thereby forming the true labels, while the other 50\% was sampled from various areas of FoS. We classified the English documents and used the document IDs to extract the German version of the same texts. A document was considered domain-related if a confidence in the positive label was larger than 0.9, and we collected the query-document pairs where the documents were labeled as domain-related. The resulting dataset contained approximately 2.27M training pairs, where 132K pairs were positive.

\end{document}